\documentclass[sigconf]{acmart}

\AtBeginDocument{%
  \providecommand\BibTeX{{%
    \normalfont B\kern-0.5em{\scshape i\kern-0.25em b}\kern-0.8em\TeX}}}

\usepackage{wrapfig}
\usepackage{multirow}
\usepackage{subfigure}
\usepackage[ruled, linesnumbered]{algorithm2e}
\usepackage{array} 
\usepackage{colortbl} 
\begin{document}

\title{MER 2024: Semi-Supervised Learning, Noise Robustness, and Open-Vocabulary Multimodal Emotion Recognition}


\author{Zheng Lian}
\affiliation{%
	\institution{Institute of Automation, Chinese Academy of Sciences (CAS)}
	\city{Beijing}
	\country{China}
}

\author{Haiyang Sun}
\affiliation{%
	\institution{Institute of Automation, CAS}
	\city{Beijing}
	\country{China}
}

\author{Licai Sun}
\affiliation{%
	\institution{Institute of Automation, CAS}
	\city{Beijing}
	\country{China}
}

\author{Zhuofan Wen}
\affiliation{%
	\institution{Institute of Automation, CAS}
	\city{Beijing}
	\country{China}
}

\author{Siyuan Zhang}
\affiliation{%
	\institution{Institute of Automation, CAS}
	\city{Beijing}
	\country{China}
}

\author{Shun Chen}
\affiliation{%
	\institution{Institute of Automation, CAS}
	\city{Beijing}
	\country{China}
}

\author{Hao Gu}
\affiliation{%
	\institution{Institute of Automation, CAS}
	\city{Beijing}
	\country{China}
}

\author{Jinming Zhao}
\affiliation{%
	\institution{Renmin University of China}
	\city{Beijing}
	\country{China}
}

\author{Ziyang Ma}
\affiliation{%
	\institution{Shanghai Jiao Tong University}
	\city{Shanghai}
	\country{China}
}

\author{Xie Chen}
\affiliation{%
	\institution{Shanghai Jiao Tong University}
	\city{Shanghai}
	\country{China}
}

\author{Jiangyan Yi}
\affiliation{%
	\institution{Institute of Automation, CAS}
	\city{Beijing}
	\country{China}
}

\author{Rui Liu}
\affiliation{%
	\institution{Inner Mongolia University}
	\city{Inner Mongolia}
	\country{China}
}

\author{Kele Xu}
\affiliation{%
	\institution{National University of Defense Technology}
	\city{Beijing}
	\country{China}
}

\author{Bin Liu}
\affiliation{%
	\institution{Institute of Automation, CAS}
	\city{Beijing}
	\country{China}
}

\author{Erik Cambria}
\affiliation{%
	\institution{Nanyang Technological University}
	\country{Singapore}
}

\author{Guoying Zhao}
\affiliation{%
	\institution{University of Oulu}
	\city{Oulu}
	\country{Finland}
}

\author{Björn W. Schuller}
\affiliation{%
	\institution{Imperial College London}
	\city{London}
	\country{United Kingdom}
}

\author{Jianhua Tao}
\affiliation{%
	\institution{Tsinghua University}
	\city{Beijing}
	\country{China}
}
\renewcommand{\shortauthors}{author name and author name, et al.}

\begin{abstract}
	Multimodal emotion recognition is an important research topic in artificial intelligence. Over the past few decades, researchers have made remarkable progress by increasing the dataset size and building more effective algorithms. However, due to problems such as complex environments and inaccurate annotations, current systems are hard to meet the demands of practical applications. Therefore, we organize the MER series of competitions to promote the development of this field. Last year, we launched MER2023\footnote{\emph{http://merchallenge.cn/mer2023}}, focusing on three interesting topics: multi-label learning, noise robustness, and semi-supervised learning. In this year's MER2024\footnote{\emph{https://zeroqiaoba.github.io/MER2024-website}}, besides expanding the dataset size, we further introduce a new track around open-vocabulary emotion recognition. The main purpose of this track is that existing datasets usually fix the label space and use majority voting to enhance the annotator consistency. However, this process may lead to inaccurate annotations, such as ignoring non-majority or non-candidate labels. In this track, we encourage participants to generate any number of labels in any category, aiming to describe emotional states as accurately as possible. Our baseline code relies on MERTools\footnote{\emph{https://github.com/zeroQiaoba/MERTools}} and is available at: \textcolor[rgb]{0.93,0.0,0.47}{https://github.com/zeroQiaoba/MERTools/tree/master/MER2024}.
\end{abstract}





\maketitle

\section{Introduction}
Multimodal emotion recognition plays an important role in human-computer interaction. Recently, researchers have made significant progress in this field. However, this task is still not well solved and its performance still cannot meet the requirements of practical applications \cite{lian2024merbench}. To this end, last year, we launched MER2023 \cite{lian2023mer}, focusing on three important topics: multi-label learning, noise robustness, and semi-supervised learning. This year, we continue the latter two tracks and introduce a new track around open-vocabulary emotion recognition, aiming to describe emotional states accurately.

First, it is hard to collect samples with emotion labels. On the one hand, the collected samples are often emotionless (i.e. \emph{neutral}) \cite{poria2019meld}. On the other hand, researchers usually hire multiple annotators and use majority voting to improve label consistency \cite{li2017reliable}, greatly increasing the annotation cost. To address the sparsity of emotional data, previous works used unlabeled data and focused on unsupervised or semi-supervised learning. Recently, MERBench \cite{lian2024merbench} has conducted a systematic analysis, pointing out the necessity of using unlabeled data from the same domain as labeled data. Therefore, we provide a large number of human-centric unlabeled videos in \textbf{MER-SEMI} and encourage participants to explore more effective unsupervised or semi-supervised strategies.

Secondly, in real scenarios, we cannot ensure that every video is free of audio noise and every frame is in high resolution. To copy with complex environments, emotion recognition systems should have a certain degree of noise robustness. Therefore, we organize \textbf{MER-NOISE} to fairly evaluate the noise robustness of different systems. Although there are many types of noise, we only consider the two most common ones: audio additive noise and image blur noise. In this track, we encourage participants to use data augmentation \cite{hazarika2022analyzing} or other effective techniques \cite{sun2023efficient, lian2023gcnet} to improve performance under noisy conditions.

Thirdly, to improve label consistency, existing datasets usually restrict the label space to a few discrete categories, then employ multiple annotators and use majority voting to select the most likely label. However, this process may cause some correct but non-candidate or non-majority labels to be ignored. Therefore, we introduce \textbf{MER-OV}, centered around open-vocabulary emotion recognition. We encourage participants to generate any number of labels in any category, trying to describe emotional states accurately.

In summary, MER2024 consists of three tracks: MER-SEMI, MER-NOISE, and MER-OV. In MER-SEMI, we encourage participants to use unlabeled data during training; in MER-NOISE, we focus on noise robustness; in MER-OV, we require participants to describe emotional states as accurately as possible. The MER series of challenges aims to provide participants with a common platform to fairly compare the performance of different techniques. In the rest of this paper, we will introduce the datasets, baselines, evaluation metrics, and experimental results in detail.

\begin{figure}[t]
	\centering
	\includegraphics[width=0.9\linewidth]{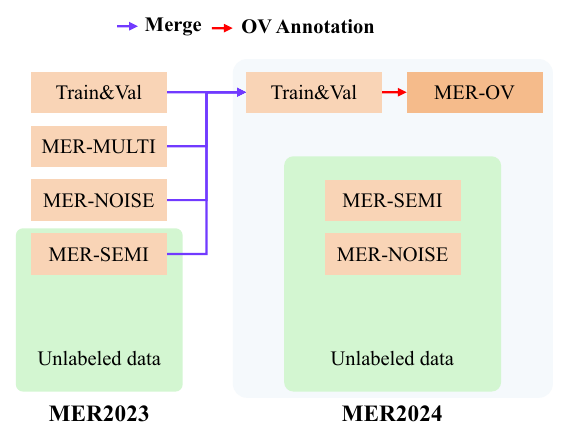}
	\caption{Dataset construction pipeline of MER2024.}
	\label{Figure1}
\end{figure}

\section{Challenge Dataset}
MER2024 is an extended version of MER2023 \cite{lian2023mer} and its construction process is summarized in Figure \ref{Figure1}. Specifically, MER2023 contains four subsets: Train$\&$Val, MER-MULTI, MER-NOISE, and MER-SEMI. In the last subset, in addition to the labeled data, it also contains a large amount of unlabeled data. In MER2024, we merge all labeled samples and obtain the updated Train$\&$Val. Meanwhile, we collect more unlabeled data and select a subset for annotation, getting MER-SEMI and MER-NOISE. For MER-OV, we select 332 samples from Train$\&$Val and provide open-vocabulary labels. Table \ref{Table1} summarizes the statistics of these datasets.

\subsection{Data Collection}
\label{sec:2-1}
As shown in Figure \ref{Figure1}, MER2024 expands the dataset size. Its data collection process is borrowed from MER2023 and includes two key steps: video cutting and video filtering.

\textbf{Video Cutting.}
The raw data in MER2024 comes from movies and TV series, which are close to real scenarios. However, these videos are usually long and have many characters, so they need to be segmented into video clips. In this process, we require that the content in these video clips is relatively complete. For videos with subtitles, the timestamps in the subtitles provide accurate boundary information, and we use them for video segmentation. For videos without subtitles, we use the voice activity detection toolkit, Silero VAD\footnote{\emph{https://github.com/snakers4/silero-vad}}, to segment the video and the speaker identification toolkit, Deep Speaker\footnote{\emph{https://github.com/philipperemy/deep-speaker}}, to merge the consecutive clips if they are likely to be from the same speaker.

\textbf{Video Filtering.}
Through the above process, we can generate video clips containing relatively complete content from the same speaker. However, it only guarantees that the audio is from the same person, but cannot guarantee that there is only one person in the visual frames. Therefore, we further filter these video clips. Specifically, we use the face detection toolkit, YuNet\footnote{\emph{https://github.com/ShiqiYu/libfacedetection}}, to ensure that most frames contain only one face. Then, we use the face recognition toolkit, face.evoLVe\footnote{\emph{https://github.com/ZhaoJ9014/face.evoLVe}}, to ensure that these faces belong to the same person. Meanwhile, the length of video clips is also important. Video clips that are too short may not convey the complete meaning; video clips that are too long may contain emotional changes, making it difficult to describe the emotional state. Therefore, we only select video clips between 2$\sim$16 seconds.

\begin{table}[t]
	\centering
	\renewcommand\tabcolsep{2pt}
	\caption{Statistics of MER2023 and MER2024. Here, ``A/B'' means that there are B samples in the subset and we only annotate A samples for performance evaluation.}
	\label{Table1}
	\begin{tabular}{lc|lc}
		\hline
		{MER2023} & {\begin{tabular}[c]{@{}c@{}}\# of samples \\ (labeled/whole)\end{tabular}} & {MER2024} & {\begin{tabular}[c]{@{}c@{}}\# of samples \\ (labeled/whole)\end{tabular}} \\
		\hline 
		Train$\&$Val 	& 3373 		& Train$\&$Val 	& 5030 \\ 
		
		MER-MULTI 		& 411 			& MER-SEMI 		& 1169/115595 \\
		MER-NOISE 		& 412 			& MER-NOISE 	& 1170/115595 \\
		MER-SEMI 		& 834/73982		& MER-OV 	& 322   \\
		\hline
	\end{tabular}
\end{table}

\begin{table*}[t]
	\centering
	\renewcommand\arraystretch{0.9}
	\caption{Model cards for some representative feature extractors. Here, ``CH'', ``EN'', and ``MULTI'' are the abbreviations of Chinese, English, and multilingualism, respectively.}
	\label{Table6}
	\begin{tabular}{l|l|l}
		\hline
		Feature & Training Data (Language) & Link \\
		\hline
		\multicolumn{3}{c}{Visual Modality} \\
		\hline
		VideoMAE-base 	& Kinetics-400 & \textcolor[rgb]{0.93,0.0,0.47}{huggingface.co/MCG-NJU/videomae-base} \\
		VideoMAE-large  	& Kinetics-400 & \textcolor[rgb]{0.93,0.0,0.47}{huggingface.co/MCG-NJU/videomae-large} \\
		CLIP-base 		& WIT & \textcolor[rgb]{0.93,0.0,0.47}{huggingface.co/openai/clip-vit-base-patch32} \\
		CLIP-large 		& WIT & \textcolor[rgb]{0.93,0.0,0.47}{huggingface.co/openai/clip-vit-large-patch14} \\
		EVA-02-base 	& ImageNet-22k & \textcolor[rgb]{0.93,0.0,0.47}{huggingface.co/timm/eva02\_base\_patch14\_224.mim\_in22k} \\
		DINOv2-large 	& LVD-142M & \textcolor[rgb]{0.93,0.0,0.47}{huggingface.co/facebook/dinov2-large} \\
		VideoMAE-base (VoxCeleb2) & VoxCeleb2 & \textcolor[rgb]{0.93,0.0,0.47}{github.com/zeroQiaoba/MERTools}\\
		VideoMAE-base (MER2023)   & MER2023 & \textcolor[rgb]{0.93,0.0,0.47}{github.com/zeroQiaoba/MERTools}\\
		
		\hline
		\multicolumn{3}{c}{Acoustic Modality} \\
		\hline
		
		emotion2vec & Mix (EN) & \textcolor[rgb]{0.93,0.0,0.47}{github.com/ddlBoJack/emotion2vec} \\
		Whisper-base 		& Internet (MULTI, mainly EN)& \textcolor[rgb]{0.93,0.0,0.47}{huggingface.co/openai/whisper-base} \\
		Whisper-large 	& Internet (MULTI, mainly EN) & \textcolor[rgb]{0.93,0.0,0.47}{huggingface.co/openai/whisper-large-v2} \\
		wav2vec 2.0-base 	& WenetSpeech (CH) & \textcolor[rgb]{0.93,0.0,0.47}{huggingface.co/TencentGameMate/chinese-wav2vec2-base} \\
		wav2vec 2.0-large 	& WenetSpeech (CH) & \textcolor[rgb]{0.93,0.0,0.47}{huggingface.co/TencentGameMate/chinese-wav2vec2-large} \\
		HUBERT-base  	& WenetSpeech (CH) & \textcolor[rgb]{0.93,0.0,0.47}{huggingface.co/TencentGameMate/chinese-hubert-base} \\
		HUBERT-large  	& WenetSpeech (CH) & \textcolor[rgb]{0.93,0.0,0.47}{huggingface.co/TencentGameMate/chinese-hubert-large} \\
		
		\hline
		\multicolumn{3}{c}{Lexical Modality} \\
		\hline
		PERT-base 		& EXT Data (CH) & \textcolor[rgb]{0.93,0.0,0.47}{huggingface.co/hfl/chinese-pert-base} \\
		PERT-large &EXT Data (CH)&\textcolor[rgb]{0.93,0.0,0.47}{huggingface.co/hfl/chinese-pert-large}\\
		LERT-base 	& EXT Data (CH) & \textcolor[rgb]{0.93,0.0,0.47}{huggingface.co/hfl/chinese-lert-base} \\
		LERT-large &EXT Data (CH)& \textcolor[rgb]{0.93,0.0,0.47}{huggingface.co/hfl/chinese-lert-large} \\
		XLNet-base 		& EXT Data (CH) & \textcolor[rgb]{0.93,0.0,0.47}{huggingface.co/hfl/chinese-xlnet-base} \\
		MacBERT-base 	& EXT Data (CH) & \textcolor[rgb]{0.93,0.0,0.47}{huggingface.co/hfl/chinese-macbert-base} \\
		MacBERT-large &EXT Data (CH)&\textcolor[rgb]{0.93,0.0,0.47}{huggingface.co/hfl/chinese-macbert-large}\\
		RoBERTa-base &EXT Data (CH)&\textcolor[rgb]{0.93,0.0,0.47}{huggingface.co/hfl/chinese-roberta-wwm-ext}\\
		RoBERTa-large 	& EXT Data (CH) & \textcolor[rgb]{0.93,0.0,0.47}{huggingface.co/hfl/chinese-roberta-wwm-ext-large} \\
		ELECTRA-base 	& EXT Data (CH) & \textcolor[rgb]{0.93,0.0,0.47}{huggingface.co/hfl/chinese-electra-180g-base-discriminator} \\
		ELECTRA-large &EXT Data (CH)& \textcolor[rgb]{0.93,0.0,0.47}{huggingface.co/hfl/chinese-electra-180g-large-discriminator}\\
		BLOOM-7B 	& ROOTS (MULTI) & \textcolor[rgb]{0.93,0.0,0.47}{huggingface.co/bigscience/bloom-7b1} \\
		MOSS-7B 		& Mix (MULTI, mainly EN and CH) & \textcolor[rgb]{0.93,0.0,0.47}{huggingface.co/fnlp/moss-base-7b} \\
		ChatGLM2-6B 	& Mix (MULTI, mainly EN and CH) & \textcolor[rgb]{0.93,0.0,0.47}{huggingface.co/THUDM/chatglm2-6b} \\
		Baichuan-13B 	& Mix (MULTI, mainly EN and CH) & \textcolor[rgb]{0.93,0.0,0.47}{huggingface.co/baichuan-inc/Baichuan-13B-Base} \\
		\hline
		
	\end{tabular}
\end{table*}

\begin{table*}[t]
	\centering
	\caption{Prompt for generating emotion-related descriptions using MLLMs (drawn from previous works \cite{lian2023explainable}).}
	\label{Table2}
	\begin{tabular}{p{3cm}|p{13cm}}
		\hline
		Models & Prompt \\
		\hline
		
		\multirow{4}{*}{Audio LLM} & As an expert in the field of emotions, please focus on the \textcolor[rgb]{0.93,0.0,0.47}{acoustic information} in the audio to discern clues related to the emotions of the individual. Please provide a detailed description and ultimately predict the emotional state of the individual. \\
		
		\hline
		
		\multirow{4}{*}{Video LLM} & As an expert in the field of emotions, please focus on the \textcolor[rgb]{0.93,0.0,0.47}{facial expressions, body movements, environment, etc.,} in the video to discern clues related to the emotions of the individual. Please provide a detailed description and ultimately predict the emotional state of the individual in the video. \\
		
		\hline
		
		\multirow{4}{*}{Audio-Video LLM} & As an expert in the field of emotions, please focus on the \textcolor[rgb]{0.93,0.0,0.47}{facial expressions, body movements, environment, acoustic information, etc.,} in the video to discern clues related to the emotions of the individual. Please provide a detailed description and ultimately predict the emotional state of the individual in the video. \\
		
		\hline
	\end{tabular}
\end{table*}

\begin{table*}[t]
	\centering
	\renewcommand\tabcolsep{2.4pt}
	\renewcommand\arraystretch{0.9}
	\caption{Unimodal results (\%). Besides five-fold cross-validation results on Train$\&$Val, we also report the results on MER-SEMI and MER-NOISE. The values in the gray columns are used for the final ranking.}
	\label{Table3}
	\begin{tabular}{l|cc|>{\columncolor{lightgray}}cc|>{\columncolor{lightgray}}cc|cc}
		\hline
		\multirow{2}{*}{Feature} & \multicolumn{2}{c|}{Train$\&$Val} & \multicolumn{2}{c|}{MER-SEMI} & \multicolumn{2}{c|}{MER-NOISE} & \multicolumn{2}{c}{Average}\\
		& WAF $(\uparrow)$ & ACC $(\uparrow)$ & WAF $(\uparrow)$ & ACC $(\uparrow)$ & WAF $(\uparrow)$ & ACC $(\uparrow)$ & WAF $(\uparrow)$ & ACC $(\uparrow)$ \\
		
		\hline
		\multicolumn{9}{c}{Visual Modality} \\
		\hline
		
		VideoMAE-base \cite{tong2022videomae} &52.86$\pm$0.23 & 53.27$\pm$0.21&47.37$\pm$0.41 & 52.04$\pm$0.16&48.53$\pm$0.39 & 51.81$\pm$0.26 & 49.59 & 52.37\\
		EmoNet \cite{kahou2016emonets} &51.76$\pm$0.31 & 53.18$\pm$0.10&52.45$\pm$0.18 & 54.94$\pm$0.15&51.22$\pm$0.21 & 53.56$\pm$0.43 & 51.81 & 53.90\\
		VideoMAE-large \cite{tong2022videomae} &57.04$\pm$0.25 & 57.49$\pm$0.30&51.05$\pm$0.39 & 55.73$\pm$0.41&50.81$\pm$0.28 & 55.22$\pm$0.62 & 52.97 & 56.15\\
		DINOv2-large \cite{oquab2023dinov2} &58.44$\pm$0.12 & 59.57$\pm$0.15&53.41$\pm$0.26 & 57.47$\pm$0.22&52.37$\pm$0.31 & 56.10$\pm$0.14 & 54.74 & 57.71\\
		SENet-FER2013 \cite{hu2018squeeze} &57.67$\pm$0.10 & 58.79$\pm$0.16&54.62$\pm$0.16 & 56.36$\pm$0.24&52.31$\pm$0.24 & 54.79$\pm$0.16 & 54.87 & 56.65\\
		ResNet-FER2013 \cite{he2016deep} &58.73$\pm$0.30 & 59.66$\pm$0.16&53.07$\pm$0.20 & 55.13$\pm$0.27&53.26$\pm$0.29 & 55.54$\pm$0.20 & 55.02 & 56.78\\
		MANet-RAFDB \cite{zhao2021learning} &59.91$\pm$0.12 & 61.10$\pm$0.11&56.59$\pm$0.31 & 58.87$\pm$0.43&55.15$\pm$0.30 & 57.49$\pm$0.23 & 57.22 & 59.15\\
		EVA-02-base \cite{fang2023eva} &61.41$\pm$0.20 & 62.28$\pm$0.22&55.25$\pm$0.21 & 58.80$\pm$0.24&55.36$\pm$0.14 & 58.14$\pm$0.15 & 57.34 & 59.74\\
		CLIP-base \cite{radford2021learning} &61.74$\pm$0.18 & 62.56$\pm$0.20&57.16$\pm$0.14 & 60.69$\pm$0.24&57.43$\pm$0.16 & 61.25$\pm$0.19 & 58.78 & 61.50\\
		VideoMAE-base (VoxCeleb2) \cite{lian2024merbench}&63.33$\pm$0.13 & 63.84$\pm$0.14&57.98$\pm$0.20 & 61.56$\pm$0.23&57.18$\pm$0.28 & 60.06$\pm$0.19 & 59.50 & 61.82\\
		VideoMAE-base (MER2023) \cite{lian2024merbench} &64.50$\pm$0.17 & 64.93$\pm$0.19&58.34$\pm$0.32 & 62.10$\pm$0.35&57.32$\pm$0.20 & 61.07$\pm$0.31 & 60.05 & 62.70\\
		CLIP-large \cite{radford2021learning} &\textcolor[rgb]{1,0,0}{\textbf{66.66$\pm$0.12 }}&\textcolor[rgb]{1,0,0}{\textbf{ 67.17$\pm$0.12}}&\textcolor[rgb]{1,0,0}{\textbf{63.27$\pm$0.29 }}&\textcolor[rgb]{1,0,0}{\textbf{ 65.45$\pm$0.30}}&\textcolor[rgb]{1,0,0}{\textbf{60.04$\pm$0.19 }}&\textcolor[rgb]{1,0,0}{\textbf{ 62.78$\pm$0.13 }}&\textcolor[rgb]{1,0,0}{\textbf{ 63.32 }}&\textcolor[rgb]{1,0,0}{\textbf{ 65.13}}\\
		
		\hline
		\multicolumn{9}{c}{Acoustic Modality} \\
		\hline
		eGeMAPS \cite{eyben2016geneva} &39.68$\pm$0.53 & 42.88$\pm$0.38&30.27$\pm$0.58 & 33.85$\pm$0.79&28.92$\pm$0.39 & 32.05$\pm$0.43 & 32.96 & 36.26\\
		VGGish \cite{hershey2017cnn} &48.60$\pm$0.10 & 50.20$\pm$0.09&45.44$\pm$0.37 & 48.52$\pm$0.23&40.70$\pm$0.31 & 43.69$\pm$0.23 & 44.91 & 47.47\\
		Whisper-base \cite{radford2023robust} &56.65$\pm$0.13 & 57.08$\pm$0.24&59.70$\pm$0.16 & 60.60$\pm$0.12&41.26$\pm$0.41 & 42.52$\pm$0.23 & 52.54 & 53.40\\
		emotion2vec \cite{ma2023emotion2vec} &56.08$\pm$0.09 & 56.48$\pm$0.03&58.98$\pm$0.33 & 59.68$\pm$0.26&45.66$\pm$0.31 & 46.97$\pm$0.31 & 53.57 & 54.38\\
		Whisper-large \cite{radford2023robust} &63.23$\pm$0.20 & 63.27$\pm$0.20&69.24$\pm$0.87 & 70.62$\pm$0.62&53.26$\pm$0.86 & 55.34$\pm$0.57 & 61.91 & 63.08\\
		wav2vec 2.0-base \cite{baevski2020wav2vec}&64.89$\pm$0.26 & 65.14$\pm$0.30&68.12$\pm$0.34 & 69.29$\pm$0.25&53.91$\pm$0.14 & 56.87$\pm$0.29 & 62.31 & 63.76\\
		wav2vec 2.0-large \cite{baevski2020wav2vec}&65.50$\pm$0.29 & 65.83$\pm$0.27&68.12$\pm$0.30 & 70.06$\pm$0.22&56.41$\pm$0.73 & 59.12$\pm$0.63 & 63.35 & 65.00\\
		HUBERT-base \cite{hsu2021hubert}&69.26$\pm$0.13 & 69.43$\pm$0.15&78.70$\pm$0.20 & 79.28$\pm$0.20&59.55$\pm$0.70 & 59.76$\pm$0.82 & 69.17 & 69.49\\
		HUBERT-large \cite{hsu2021hubert}&\textcolor[rgb]{1,0,0}{\textbf{73.02$\pm$0.13 }}&\textcolor[rgb]{1,0,0}{\textbf{ 73.10$\pm$0.15}}&\textcolor[rgb]{1,0,0}{\textbf{83.42$\pm$0.37 }}&\textcolor[rgb]{1,0,0}{\textbf{ 84.00$\pm$0.32}}&\textcolor[rgb]{1,0,0}{\textbf{73.21$\pm$0.36 }}&\textcolor[rgb]{1,0,0}{\textbf{ 74.03$\pm$0.31 }}&\textcolor[rgb]{1,0,0}{\textbf{ 76.55 }}&\textcolor[rgb]{1,0,0}{\textbf{ 77.05}}\\
		
		\hline
		\multicolumn{9}{c}{Lexical Modality} \\
		\hline
		
		XLNet-base \cite{yang2019xlnet} &48.59$\pm$0.30 & 48.95$\pm$0.29&49.13$\pm$0.07 & 48.64$\pm$0.04&46.14$\pm$0.16 & 46.44$\pm$0.10 & 47.96 & 48.01\\
		ELECTRA-large \cite{clark2020electra} &50.45$\pm$0.15 & 50.83$\pm$0.09&52.07$\pm$0.16 & 51.81$\pm$0.12&47.34$\pm$0.18 & 47.14$\pm$0.12 & 49.95 & 49.93\\
		MOSS-7B&51.25$\pm$0.23 & 51.64$\pm$0.19&51.62$\pm$0.22 & 51.64$\pm$0.14&47.87$\pm$0.19 & 48.42$\pm$0.29 & 50.25 & 50.57\\
		PERT-large \cite{cui2022pert} &50.36$\pm$0.23 & 50.69$\pm$0.10&53.47$\pm$0.17 & 53.26$\pm$0.13&48.76$\pm$0.23 & 48.83$\pm$0.22 & 50.86 & 50.93\\
		PERT-base \cite{cui2022pert} &50.26$\pm$0.08 & 50.62$\pm$0.12&53.64$\pm$0.21 & 53.50$\pm$0.18&49.29$\pm$0.18 & 49.44$\pm$0.18 & 51.06 & 51.19\\
		LERT-large \cite{cui2022lert}&52.22$\pm$0.18 & 52.38$\pm$0.19&52.40$\pm$0.20 & 52.18$\pm$0.28&49.17$\pm$0.44 & 49.14$\pm$0.49 & 51.26 & 51.24\\
		ELECTRA-base \cite{clark2020electra}&50.98$\pm$0.07 & 51.30$\pm$0.07&54.50$\pm$0.33 & 54.35$\pm$0.30&49.35$\pm$0.15 & 49.32$\pm$0.16 & 51.61 & 51.65\\
		LERT-base \cite{cui2022lert}&52.37$\pm$0.13 & 52.72$\pm$0.10&54.84$\pm$0.21 & 54.62$\pm$0.20&47.98$\pm$0.07 & 48.15$\pm$0.15 & 51.73 & 51.83\\
		RoBERTa-base \cite{liu2019roberta} &51.84$\pm$0.22 & 52.24$\pm$0.18&54.37$\pm$0.24 & 53.99$\pm$0.34&49.49$\pm$0.51 & 49.56$\pm$0.41 & 51.90 & 51.93\\
		MacBERT-base \cite{cui2020revisiting} &51.40$\pm$0.13 & 51.75$\pm$0.13&55.11$\pm$0.29 & 54.74$\pm$0.37&49.68$\pm$0.24 & 49.74$\pm$0.25 & 52.06 & 52.08\\
		RoBERTa-large \cite{liu2019roberta} &52.66$\pm$0.18 & 52.92$\pm$0.13&55.14$\pm$0.31 & 54.88$\pm$0.31&49.06$\pm$0.21 & 49.23$\pm$0.28 & 52.29 & 52.34\\
		ChatGLM2-6B \cite{du2022glm} &53.04$\pm$0.23 & 53.28$\pm$0.22&55.52$\pm$0.56 & 55.06$\pm$0.63&50.39$\pm$0.43 & 50.40$\pm$0.40 & 52.98 & 52.91\\
		MacBERT-large \cite{cui2020revisiting} &52.19$\pm$0.14 & 52.47$\pm$0.12&\textcolor[rgb]{1,0,0}{\textbf{56.99$\pm$0.20 }}&\textcolor[rgb]{1,0,0}{\textbf{ 56.81$\pm$0.31}}&50.24$\pm$0.20 & 50.34$\pm$0.23 & 53.14 & 53.21\\
		BLOOM-7B \cite{workshop2022bloom} &53.13$\pm$0.24 & 53.30$\pm$0.24&56.01$\pm$0.36 & 55.99$\pm$0.40&50.38$\pm$0.21 & 50.58$\pm$0.13 & 53.18 & 53.29\\
		Baichuan-13B \cite{yang2023baichuan} &\textcolor[rgb]{1,0,0}{\textbf{54.86$\pm$0.12 }}&\textcolor[rgb]{1,0,0}{\textbf{ 55.11$\pm$0.06}}&56.63$\pm$0.18 & 56.17$\pm$0.34&\textcolor[rgb]{1,0,0}{\textbf{52.01$\pm$0.30 }}&\textcolor[rgb]{1,0,0}{\textbf{ 51.90$\pm$0.21 }}&\textcolor[rgb]{1,0,0}{\textbf{ 54.50 }}&\textcolor[rgb]{1,0,0}{\textbf{ 54.39}}\\
		
		\hline
	\end{tabular}
\end{table*}

\begin{table*}[t]
	\centering
	\renewcommand\tabcolsep{6pt}
	\renewcommand\arraystretch{0.9}
	\caption{Multimodal results (\%). ``A'', ``V'', and ``T'' represent acoustic, visual, and textual modalities, respectively. ``TopN'' means that we select the top-N features for each modality and their ranking is based on the average WAF in Table \ref{Table3}.}
	\label{Table4}
	\begin{tabular}{l|cc|>{\columncolor{lightgray}}cc|>{\columncolor{lightgray}}cc|cc}
		\hline
		\multirow{2}{*}{\# Top} & \multicolumn{2}{c|}{Train$\&$Val} & \multicolumn{2}{c|}{MER-SEMI} & \multicolumn{2}{c|}{MER-NOISE} & \multicolumn{2}{c}{Average}\\
		& WAF $(\uparrow)$ & ACC $(\uparrow)$ & WAF $(\uparrow)$ & ACC $(\uparrow)$ & WAF $(\uparrow)$ & ACC $(\uparrow)$ & WAF $(\uparrow)$ & ACC $(\uparrow)$ \\
		
		\hline
		\multicolumn{9}{c}{A+V} \\
		\hline
		
		Top1&78.86$\pm$0.17 & 78.98$\pm$0.13&84.07$\pm$0.17 & 84.60$\pm$0.16&\textcolor[rgb]{1,0,0}{\textbf{78.14$\pm$0.31 }}&\textcolor[rgb]{1,0,0}{\textbf{ 79.43$\pm$0.22 }}&\textcolor[rgb]{1,0,0}{\textbf{ 80.36 }}&\textcolor[rgb]{1,0,0}{\textbf{ 81.00}}\\
		Top2&78.92$\pm$0.04 & 78.97$\pm$0.11&84.20$\pm$0.15 & 84.73$\pm$0.13&77.33$\pm$0.14 & 78.55$\pm$0.18 & 80.15 & 80.75\\
		Top3&79.16$\pm$0.11 & 79.18$\pm$0.16&84.09$\pm$0.15 & 84.67$\pm$0.15&77.26$\pm$0.38 & 78.70$\pm$0.33 & 80.17 & 80.85\\
		Top4&79.11$\pm$0.08 & 79.23$\pm$0.09&84.12$\pm$0.11 & 84.72$\pm$0.11&77.16$\pm$0.15 & 78.55$\pm$0.15 & 80.13 & 80.83\\
		Top5&\textcolor[rgb]{1,0,0}{\textbf{79.17$\pm$0.07 }}& 79.25$\pm$0.10&\textcolor[rgb]{1,0,0}{\textbf{84.38$\pm$0.23 }}&\textcolor[rgb]{1,0,0}{\textbf{ 84.92$\pm$0.22}}&77.05$\pm$0.17 & 78.46$\pm$0.32 & 80.20 & 80.88\\
		Top6&79.12$\pm$0.09 &\textcolor[rgb]{1,0,0}{\textbf{ 79.25$\pm$0.12}}&84.14$\pm$0.20 & 84.72$\pm$0.19&77.54$\pm$0.34 & 78.78$\pm$0.22 & 80.27 & 80.91\\
		
		\hline
		\multicolumn{9}{c}{A+T} \\
		\hline
		
		Top1&73.67$\pm$0.13 & 73.82$\pm$0.09&84.06$\pm$0.35 & 84.57$\pm$0.32&\textcolor[rgb]{1,0,0}{\textbf{74.74$\pm$0.26 }}&\textcolor[rgb]{1,0,0}{\textbf{ 75.61$\pm$0.23 }}&\textcolor[rgb]{1,0,0}{\textbf{ 77.49 }}&\textcolor[rgb]{1,0,0}{\textbf{ 78.00}}\\
		Top2&73.75$\pm$0.19 & 73.84$\pm$0.17&84.50$\pm$0.36 & 85.01$\pm$0.35&73.38$\pm$0.46 & 74.25$\pm$0.59 & 77.21 & 77.70\\
		Top3&\textcolor[rgb]{1,0,0}{\textbf{73.97$\pm$0.11 }}&\textcolor[rgb]{1,0,0}{\textbf{ 73.97$\pm$0.05}}&\textcolor[rgb]{1,0,0}{\textbf{84.64$\pm$0.16 }}&\textcolor[rgb]{1,0,0}{\textbf{ 85.10$\pm$0.17}}&73.67$\pm$0.39 & 74.32$\pm$0.39 & 77.42 & 77.80\\
		Top4&73.82$\pm$0.22 & 73.87$\pm$0.22&84.37$\pm$0.57 & 84.85$\pm$0.54&73.90$\pm$0.30 & 74.36$\pm$0.29 & 77.36 & 77.69\\
		Top5&73.73$\pm$0.19 & 73.80$\pm$0.18&84.08$\pm$0.30 & 84.64$\pm$0.27&73.53$\pm$0.14 & 74.20$\pm$0.15 & 77.11 & 77.55\\
		Top6&73.68$\pm$0.12 & 73.70$\pm$0.11&84.31$\pm$0.36 & 84.80$\pm$0.31&73.99$\pm$0.49 & 74.62$\pm$0.40 & 77.32 & 77.71\\
		
		\hline
		\multicolumn{9}{c}{V+T} \\
		\hline
		
		Top1&72.28$\pm$0.16 & 72.47$\pm$0.11&77.34$\pm$0.30 & 77.74$\pm$0.28&71.98$\pm$0.42 & 73.38$\pm$0.31 & 73.87 & 74.53\\
		Top2&73.88$\pm$0.08 & 74.03$\pm$0.07&77.17$\pm$0.43 & 77.89$\pm$0.39&71.99$\pm$0.21 & 73.72$\pm$0.11 & 74.34 & 75.21\\
		Top3&74.29$\pm$0.09 & 74.35$\pm$0.10&77.98$\pm$0.37 & 78.60$\pm$0.39&72.61$\pm$0.34 & 74.26$\pm$0.35 & 74.96 & 75.74\\
		Top4&74.06$\pm$0.12 & 74.22$\pm$0.09&\textcolor[rgb]{1,0,0}{\textbf{78.31$\pm$0.34 }}&\textcolor[rgb]{1,0,0}{\textbf{ 79.08$\pm$0.33}}&72.51$\pm$0.18 & 74.26$\pm$0.13 & 74.96 & 75.85\\
		Top5&74.04$\pm$0.10 & 74.20$\pm$0.10&77.78$\pm$0.23 & 78.50$\pm$0.25&\textcolor[rgb]{1,0,0}{\textbf{73.12$\pm$0.41 }}&\textcolor[rgb]{1,0,0}{\textbf{ 74.81$\pm$0.33 }}& 74.98 & 75.84\\
		Top6&\textcolor[rgb]{1,0,0}{\textbf{74.33$\pm$0.17 }}&\textcolor[rgb]{1,0,0}{\textbf{ 74.48$\pm$0.21}}&77.83$\pm$0.26 & 78.53$\pm$0.26&73.03$\pm$0.36 & 74.70$\pm$0.27 &\textcolor[rgb]{1,0,0}{\textbf{ 75.07 }}&\textcolor[rgb]{1,0,0}{\textbf{ 75.90}}\\

		\hline
		\multicolumn{9}{c}{A+V+T} \\
		\hline
		
		Top1&79.31$\pm$0.03 & 79.40$\pm$0.04&\textcolor[rgb]{1,0,0}{\textbf{86.73$\pm$0.13 }}&\textcolor[rgb]{1,0,0}{\textbf{ 87.09$\pm$0.16}}&79.47$\pm$0.29 & 80.67$\pm$0.19 & 81.84 & 82.39\\
		Top2&80.05$\pm$0.06 & 80.07$\pm$0.09&85.94$\pm$0.25 & 86.44$\pm$0.24&79.11$\pm$0.33 & 80.33$\pm$0.29 & 81.70 & 82.28\\
		Top3&80.28$\pm$0.16 & 80.33$\pm$0.12&86.27$\pm$0.22 & 86.74$\pm$0.18&78.75$\pm$0.20 & 80.17$\pm$0.19 & 81.77 & 82.41\\
		Top4&80.02$\pm$0.11 & 80.10$\pm$0.11&86.19$\pm$0.14 & 86.67$\pm$0.16&\textcolor[rgb]{1,0,0}{\textbf{79.62$\pm$0.27 }}&\textcolor[rgb]{1,0,0}{\textbf{ 80.82$\pm$0.28 }}& 81.94 & 82.53\\
		Top5&80.17$\pm$0.13 & 80.21$\pm$0.12&85.93$\pm$0.15 & 86.45$\pm$0.17&78.91$\pm$0.28 & 80.36$\pm$0.22 & 81.67 & 82.34\\
		Top6&\textcolor[rgb]{1,0,0}{\textbf{80.34$\pm$0.10 }}&\textcolor[rgb]{1,0,0}{\textbf{ 80.43$\pm$0.12}}&86.32$\pm$0.22 & 86.86$\pm$0.19&79.24$\pm$0.40 & 80.33$\pm$0.28 &\textcolor[rgb]{1,0,0}{\textbf{ 81.97 }}&\textcolor[rgb]{1,0,0}{\textbf{ 82.54}}\\
		
		\hline
	\end{tabular}
\end{table*}

\subsection{MER-SEMI and MER-NOISE}
Annotating all unlabeled data requires a lot of labor costs. To reduce the cost, we only select samples with a high probability of explicit emotions. Specifically, we use the top-16 results in last year's challenge and calculate the proportion of primary labels as the basis for selection. For example, if a sample is predicted as \emph{happy} 10 times and \emph{neutral} 6 times, its score is $v=\max(6, 10)/16$. Then, taking into account the calculated score and class balance, we select a total of 6,000 samples for labeling. During the annotation process, we hire 5 annotators and only select samples in which at least 4 annotators assign the same label, resulting in 2,339 labeled samples. All these samples are equally divided into two parts, one for MER-SEMI and the other for MER-NOISE.

For MER-NOISE, we additionally add noise to videos. This paper considers two types of noise: audio additive noise and image blur noise. For the audio, we select noise from the MUSAN dataset \cite{snyder2015musan}, which contains three subsets: \emph{music}, \emph{noise}, and \emph{speech}. The noise in the first two subsets may convey emotions and affect the emotion of the raw audio. For example, rain and thunder may lead to a negative sentiment, while a pleasant breeze may generate a positive sentiment. Therefore, we only use the noise in the last subset. Specifically, we randomly select a speech-to-noise ratio (SNR) between 5dB$\sim$10dB and randomly select noise from the \emph{speech} subset. For the video, image blur is a common noise. To generate blurry images, we downsample the image to lose some details and then upsample the low-resolution image to keep the size unchanged. This paper randomly selects a downsampling factor from $r=\{1, 2, 4\}$.

\subsection{MER-OV}
\label{sec:2-3}
Unlike MER-SEMI and MER-NOISE which predict the most likely emotion within a fixed label space, MER-OV needs to predict any number of emotions in any category. Previously, researchers made an initial attempt at this task \cite{lian2023explainable}. They first annotated emotion-related acoustic and visual clues. Then, they relied on the reasoning ability of LLMs to disambiguate subtitles using these clues. This process can generate descriptions with rich emotions. After that, they used GPT-3.5 (``gpt-3.5-turbo-16k-0613'') to extract all labels using the following prompt: \emph{\textcolor[rgb]{0.93,0.0,0.47}{Please assume the role of an expert in the field of emotions. We provide clues that may be related to the emotions of the characters. Based on the provided clues, please identify the emotional states of the main characters. Please separate different emotional categories with commas and output only the clearly identifiable emotional categories in a list format. If none are identified, please output an empty list.}}

Finally, they performed the manual check and got the ground truth $Y_{gt}$. Through the above process, each sample can have an average of 3 labels. However, due to the high cost, they only annotated 332 samples \cite{lian2023explainable}. For MER-OV, participants can borrow some basic ideas from this process. Meanwhile, we encourage participants to use MLLMs or to further conduct instruction fine-tuning.

\subsection{Challenge Protocol}
To download the dataset, participants should fill out an End User License Agreement (EULA)\footnote{\emph{https://drive.google.com/file/d/1cXNfKHyJzVXg\_7kWSf\_nVKtsxIZVa517/view?usp=sharing}}. It asks participants to use this dataset only for academic research and not to edit or upload it to the Internet. For MER-SEMI and MER-NOISE, each team should predict the most likely label among 6 categories (i.e., \emph{worried}, \emph{happy}, \emph{neutral}, \emph{angry}, \emph{surprised}, and \emph{sad}). For MER-OV, each team can submit any number of labels in any category. Meanwhile, participants cannot use closed-source models (such as GPT or Claude) in MER-OV. For all tracks, participants should predict results for 115,595 unlabeled data, although we only evaluate a small subset of them. It requires participants to focus on the generalization ability and develop systems that do not require a lot of inference time. Additionally, participants cannot manually annotate samples in MER2024. We will ask them to submit the code for further checking. Finally, each team should submit a paper describing their method. For other requirements, please refer to our official website\footnote{\emph{https://zeroqiaoba.github.io/MER2024-website}}.

\section{Baselines and Evaluation Metrics}
In this section, we first introduce the baselines of three tracks. Then, we illustrate implementation details and evaluation metrics.

\subsection{MER-SEMI and MER-NOISE}
For MER-SEMI and MER-NOISE, we build baselines based on MERTools\footnote{\emph{https://github.com/zeroQiaoba/MERTools}}. Feature selection and fusion are crucial for emotion recognition systems. For feature selection, we adopt the recommendations of MERBench \cite{lian2024merbench}. Language compatibility is important for lexical and acoustic features, so we mainly select encoders trained on Chinese corpora. Domain compatibility is important for visual features. Therefore, besides encoders trained on action or caption datasets, we also select encoders trained on human-centric videos. Table \ref{Table6} lists the model cards of some representative encoders.

Regarding the fusion strategy, MERBench points out that the attention mechanism can achieve relatively good performance \cite{lian2024merbench}. The reason lies in that the labeled samples are limited in emotion recognition. Complex architectures may cause overfitting problems, which will affect the model's generalization ability on unseen data. Assume that acoustic, lexical, and visual features are $f_a \in \mathbb{R}^{d_a}$, $f_l \in \mathbb{R}^{d_l}$, $f_v \in \mathbb{R}^{d_v}$, respectively. During the fusion process, we first map them to a fixed dimension $d$:
\begin{equation}
h_m = \mbox{ReLU}\left(f_mW_m+b_m\right), m \in \{a, l, v\},
\end{equation}
where $W_m \in \mathbb{R}^{d_m \times d}$ and $b_m \in \mathbb{R}^{d}$ are trainable parameters. Then, we calculate the attention score for each modality:
\begin{equation}
h = \mbox{Concat}\left(h_a, h_l, h_v\right),
\end{equation}
\begin{equation}
\alpha = \mbox{softmax}\left(h^TW_\alpha+b_\alpha\right),
\end{equation}
where $W_\alpha \in \mathbb{R}^{d \times 1}$ and $b_\alpha \in \mathbb{R}^{1}$ are trainable parameters. Here, $h \in \mathbb{R}^{d \times 3}$ and $\alpha \in \mathbb{R}^{3 \times 1}$. Finally, the fused multimodal features $z=h \alpha \in \mathbb{R}^{d}$ are used for emotion recognition.

\subsection{MER-OV}
OV emotion recognition is a new task and the lack of large-scale datasets makes it difficult to conduct supervised training. Therefore, we choose pre-trained MLLMs as baselines because they can handle various multimodal tasks without further training. To solve OV emotion recognition using MLLMs, we borrow the dataset construction pipeline in Section \ref{sec:2-3}. Specifically, we first use MLLMs and the prompts in Table \ref{Table2} to extract multifaceted and multimodal emotion-related clues. Then, we use these clues to disambiguate the subtitle and generate descriptions with rich emotions. Finally, we extract all emotion labels using the prompt in Section \ref{sec:2-3}.

\subsection{Implementation Details}
For MER-SEMI and MER-NOISE, we use the attention mechanism for multimodal fusion. This process involves one hyper-parameter, the dimension of the hidden representation $d$, and we choose it from $\{64, 128, 256\}$. During training, we use the Adam optimizer and choose the learning rate from $\{10^{-3}, 10^{-4}\}$. To alleviate the overfitting problem, we also use Dropout and select the rate from $\{0.2, 0.3, 0.4, 0.5\}$. Therefore, a total of 3 hyper-parameters need to be adjusted. To find the optimal hyper-parameters, we randomly select 50 parameter combinations and choose the best-performing combination in five-fold cross-validation. Finally, we report its average result and standard deviation.

For MER-OV, we directly use pretrained MLLMs. Due to limited GPU memory, we use their 7B weights by default. All models are implemented with PyTorch and all inference processes are carried out with a 32G NVIDIA Tesla V100 GPU.

\subsection{Evaluation Metrics}
For MER-SEMI and MER-NOISE, we choose two widely used metrics in emotion recognition for performance evaluation: accuracy and weighted average f1-score (WAF) \cite{lian2021ctnet}. Considering the inherent class imbalance, we choose WAF for the final ranking. 

For MER-OV, we use set-level accuracy and recall for performance evaluation, consistent with previous works \cite{lian2023explainable}. Specifically, assume that the true label set is $Y_{gt}=\{y_i\}_{i=1}^{M}$ and the predicted label set is $\hat{Y}_{p}=\{\hat{y}_i\}_{i=1}^{N}$, where $M$ and $N$ are the number of labels. Since we do not fix the label space, there may be synonyms, i.e., labels with different expressions but the same meaning. Therefore, we first group all labels using GPT-3.5: \emph{\textcolor[rgb]{0.93,0.0,0.47}{Please assume the role of an expert in the field of emotions. We provide a set of emotions. Please group the emotions, with each group containing emotions with the same meaning. Directly output the results. The output format should be a list containing multiple lists.}} Next, we use the GPT-based grouping results $G(\cdot)$ to map all labels to their group IDs:
\begin{equation}
{Y}_{gt}^{m} = \{G(x) |x \in \{{y}_i\}_{i=1}^M\}, \hat{Y}_p^{m} = \{G(x) |x \in \{\hat{y}_i\}_{i=1}^N\}.
\end{equation}
We then calculate set-level accuracy and recall and use their average for the final ranking, consistent with previous works \cite{lian2023explainable}.
\begin{equation}
\mbox{Accuracy}_{{s}} = \frac{|{Y}_{gt}^{m} \cap \hat{{Y}}_p^{m}|}{|\hat{{Y}}_p^{m}|}, \mbox{Recall}_{{s}} = \frac{|{Y}_{gt}^{m} \cap \hat{{Y}}_p^{m}|}{|{Y}_{gt}^{m}|}.
\end{equation}
\begin{equation}
\mbox{Avg} = \frac{\mbox{Accuracy}_{{s}} + \mbox{Recall}_{{s}}}{2}.
\end{equation}

\section{Results and Discussion}
This section reports baseline results for three tracks. For MER-SEMI and MER-NOISE, we report unimodal and multimodal results. For MER-OV, we report the results of various MLLMs.

\subsection{MER-SEMI and MER-NOISE}
Table \ref{Table3} shows the unimodal results. From this table, we observe that models that perform well on Train$\&$Val generally perform well on MER-SEMI and MER-NOISE. These results suggest that the models trained on our dataset have a good generalization ability.

For the visual modality, unsupervised or semi-supervised models (e.g., CLIP-large) generally outperform supervised models (e.g., SENet-FER2013). This phenomenon suggests that unsupervised or semi-supervised strategies can capture universal representations, which are also helpful for emotion recognition. Meanwhile, previous works have emphasized the importance of domain compatibility \cite{lian2024merbench}. Therefore, we take VideoMAE-base as an example and further train it on in-domain corpora, such as VoxCeleb2 and MER2023. This process leads to significant performance improvements compared to the original model. Consequently, we recommend participants train other visual encoders on in-domain corpora.

For the acoustic and lexical modalities, we primarily choose encoders trained on the Chinese corpus, as suggested by MERBench \cite{lian2024merbench}. For the acoustic modality, we observe that unsupervised or semi-supervised models (e.g., HUBERT-large) generally perform better than traditional encoders (e.g., eGeMAPS), which is consistent with the findings in the visual modality. For the lexical modality, we observe that LLMs generally outperform pretrained language models (PLMs). This suggests that by increasing the training data and model size, we can build more powerful lexical encoders.

Table \ref{Table4} shows the multimodal results. In this table, we choose the top-N features for each modality and use the attention mechanism for multimodal fusion. Their ranking is based on the average WAF in Table \ref{Table3}. We observe that different modality combinations prefer distinct $N$. Therefore, we should adjust it for each combination. Meanwhile, the trimodal results generally perform best, reflecting the effectiveness of multimodal fusion and the necessity of each modality. Overall, in terms of WAF scores, our baseline system can reach 86.73\% in MER-SEMI and 79.62\% in MER-NOISE. 

\begin{table}[t]
	\centering
	\renewcommand\tabcolsep{2pt}
	\caption{Baseline results on MER-OV (taken from previous works \cite{lian2023explainable}). The values in the gray column are used for the final ranking. Here, ``L'', ``V'', and ``A'' indicate whether lexical, visual, and acoustic information are used during inference.}
	\label{Table5}
	\begin{tabular}{l|ccc|>{\columncolor{lightgray}}ccc}
		\hline
		Model & L & V & A & Avg & $\mbox{Accuracy}_{\mbox{s}}$ & $\mbox{Recall}_{\mbox{s}}$ \\
		\hline
		Empty          & $\times$ & $\times$ & $\times$&0.00$\pm$0.00 & 0.00$\pm$0.00 & 0.00$\pm$0.00  \\
		Random         & $\times$ & $\times$ & $\times$&19.13$\pm$0.06 & 24.85$\pm$0.15 & 13.42$\pm$0.04  \\
		\hline
		Qwen-Audio \cite{chu2023qwen}   & $\surd$  & $\times$ & $\surd$&40.23$\pm$0.09 & 49.42$\pm$0.18 & 31.04$\pm$0.00  \\
		OneLLM \cite{han2023onellm}     & $\surd$  & $\times$ & $\surd$&43.04$\pm$0.06 & 45.92$\pm$0.05 & 40.15$\pm$0.06  \\
		Otter  \cite{li2023otter}   & $\surd$  & $\surd$ & $\times$&44.40$\pm$0.09 & 50.71$\pm$0.10 & 38.09$\pm$0.09 \\
		VideoChat  \cite{li2023videochat} & $\surd$  & $\surd$ & $\times$&45.70$\pm$0.09 & 42.90$\pm$0.27 & 48.49$\pm$0.10  \\
		Video-LLaMA \cite{zhang2023video}  & $\surd$ & $\surd$ & $\times$&44.74$\pm$0.14 & 44.14$\pm$0.13 & 45.34$\pm$0.15  \\
		PandaGPT  \cite{su2023pandagpt}     & $\surd$  & $\surd$  & $\surd$&46.21$\pm$0.17 & 50.03$\pm$0.01 & 42.38$\pm$0.33 \\
		SALMONN \cite{tang2023salmonn}      & $\surd$  & $\times$ & $\surd$&48.06$\pm$0.04 & 50.20$\pm$0.04 & 45.92$\pm$0.04 \\
		Video-LLaVA \cite{lin2023video}   & $\surd$  & $\surd$ & $\times$&47.12$\pm$0.15 & 48.58$\pm$0.02 & 45.66$\pm$0.29 \\
		VideoChat2 \cite{li2024mvbench}   & $\surd$  & $\surd$ & $\times$&49.60$\pm$0.28 & 54.72$\pm$0.41 & 44.47$\pm$0.15  \\
		OneLLM \cite{han2023onellm}     & $\surd$  & $\surd$  & $\times$&50.99$\pm$0.08 & 55.93$\pm$0.09 & 46.06$\pm$0.06  \\
		LLaMA-VID \cite{li2023llama}    & $\surd$  & $\surd$  & $\times$&51.29$\pm$0.09 & 52.71$\pm$0.18 & 49.87$\pm$0.00  \\
		mPLUG-Owl \cite{ye2023mplug}   & $\surd$  & $\surd$ & $\times$&52.79$\pm$0.13 & 54.54$\pm$0.13 & 51.04$\pm$0.13  \\
		Video-ChatGPT \cite{maaz2023video} & $\surd$ & $\surd$ & $\times$&50.73$\pm$0.06 & 54.03$\pm$0.04 & 47.44$\pm$0.07 \\
		Chat-UniVi \cite{jin2023chat}   & $\surd$  & $\surd$  & $\times$&53.09$\pm$0.01 & 53.68$\pm$0.00 & 52.50$\pm$0.02  \\
		GPT-4V \cite{openai2023gpt4v}     & $\surd$   & $\surd$ & $\times$&56.69$\pm$0.04 & 48.52$\pm$0.07 & 64.86$\pm$0.00  \\
		\hline
	\end{tabular}
\end{table}

\subsection{MER-OV}
This section discusses the performance of different methods on MER-OV. In addition to MLLMs, we introduce two heuristic baselines: \emph{Empty} and \emph{Random}. For the former, we do not assign any label. For the latter, we randomly select a label from six candidate categories (i.e., \emph{worried}, \emph{happy}, \emph{neutral}, \emph{angry}, \emph{surprised}, and \emph{sad}). Experimental results are shown in Table \ref{Table5}. We observe that MLLMs generally outperform heuristic baselines, indicating that they can solve this task to some extent. However, there is still a significant gap between MLLMs and ground truth, indicating the difficulty of this task. We recommend participants test other MLLMs or use supervised fine-tuning, which may bring performance improvement.

\section{Conclusions}
This paper introduces MER2024, an extension of the MER2023 competition. Besides including more samples, we also introduce a new track called open-vocabulary emotion recognition, requiring participants to predict any number of labels in any category, aiming to achieve more accurate emotion recognition. In this paper, we introduce the datasets, baselines, evaluation metrics, and experimental results. We also open-source the code to ensure reproducibility. We hope that the MER series of challenges can provide researchers with a common platform to fairly evaluate their emotion recognition systems and further promote the development of this field.


\bibliographystyle{unsrt}
\bibliography{mybib}

\end{document}